\documentclass[10pt, conference]{IEEEtran}
\usepackage{filecontents}

 \usepackage{tikz}
 \usepackage{textcomp}
 \usepackage{hyperref}
 \usepackage{lipsum}

\usepackage{textcomp}
\usepackage{hyperref}
\usepackage{lipsum}

\usepackage{algpseudocode}

\usepackage{spconf,amsmath,graphicx}

\title{On the Behavior of the Expectation-Maximization Algorithm for Mixture Models}

\name{Babak Barazandeh$^{\star }$ \qquad Meisam Razaviyayn$^{\dagger}$ }

\address{ $^{\star \dagger}$University of Southern California\\Email : \{$^{\star}$ barazand, $^{\dagger}$ razaviya\}@usc.edu}
\usepackage{cite}
\newcommand*\diff{\mathop{}\!\mathrm{d}}

\usepackage{amsmath}
\usepackage{ntheorem}
\usepackage{mathtools}
\usepackage{cite}
\newcommand\norm[1]{\left\lVert#1\right\rVert}

\usepackage{amsmath,amssymb,amsfonts}
\usepackage{graphicx}
\usepackage{textcomp}
\usepackage{float}
\usepackage[utf8]{inputenc} 
\usepackage[T1]{fontenc}    
\usepackage{nth}
\usepackage{hyperref}       
\usepackage{url}            
\usepackage{booktabs}       
\usepackage{amsfonts}       
\usepackage{nicefrac}       
\usepackage{microtype}      
\usepackage{subcaption}
\usepackage{amsbsy} 
\newtheorem{theorem}{Theorem}
\newtheorem*{proof*}{Proof}
\newtheorem{lemma}{Lemma}

\usepackage[linesnumbered,ruled]{algorithm2e}
 
\newcommand{\bmu}{\boldsymbol{\mu}}
\newcommand{\bx}{\mathbf{x}}
 
\newcommand*{\QED}{\hfill\ensuremath{\blacksquare}}%

\def\BibTeX{{\rm B\kern-.05em{\sc i\kern-.025em b}\kern-.08em
		T\kern-.1667em\lower.7ex\hbox{E}\kern-.125emX}}

\usepackage{eso-pic}
\AddToShipoutPicture*{\footnotesize \sffamily\raisebox{1cm}{\hspace{1cm} \parbox{\textwidth}{Copyright 2018 IEEE. Published in the IEEE 2018 Global Conference on Signal and Information Processing (GlobalSIP 2018), scheduled for November 26-28, 2018 in Anaheim, California, USA. Personal use of this material is permitted. However, permission to reprint/republish this material for advertising or promotional purposes or for creating new collective works for resale or redistribution to servers or lists, or to reuse any copyrighted component of this work in other works, must be obtained from the IEEE. Contact: Manager, Copyrights and Permissions / IEEE Service Center / 445 Hoes Lane / P.O. Box 1331 / Piscataway, NJ 08855-1331, USA. Telephone: + Intl. 908-562-3966}}}

\begin{document}

		\setlength{\abovedisplayskip}{1.5pt}
	\setlength{\belowdisplayskip}{1.5pt}
%

\maketitle

\begin{abstract}

Finite mixture models are among the most popular statistical models used in different data science disciplines. Despite their broad applicability, inference under these models typically leads to computationally challenging non-convex problems. While the Expectation-Maximization (EM) algorithm is the most popular approach for solving these non-convex problems, the behavior of this algorithm is not well understood. In this work, we focus on the case of mixture of Laplacian (or Gaussian) distribution. We start by analyzing a simple equally weighted mixture of two single dimensional Laplacian distributions and show that every local optimum of the population maximum likelihood estimation problem is globally optimal. Then, we prove that the EM algorithm converges to the ground truth parameters almost surely with random initialization. Our result extends the existing  results for Gaussian distribution to Laplacian distribution.  Then we numerically study the behavior of  mixture models with  more than two components. Motivated by our extensive numerical experiments, we propose a novel stochastic method for estimating the mean of components of a mixture model.  Our numerical experiments show that our algorithm outperforms the Na\"{i}ve EM algorithm in almost all scenarios.

\end{abstract}
\section{Introduction}
The ability of finite mixture distributions \cite{pearson1894contributions} to model the presence of subpopulations within an overall population has made them popular across almost all engineering and scientific disciplines \cite{melnykov2010finite,zhang2015finite,titterington1985statistical,mclachlan2004finite}. While statistical identifiability for various mixture models has been widely studied \cite{teicher1963identifiability, allman2009identifiability}, Gaussian mixture model (GMM) has drawn more attention due to its wide applicability \cite{day1969estimating,wolfe1970pattern}. Started by Dasgupta\cite{dasgupta1999learning}, there have been multiple efforts for finding algorithm with polynomial sample/time complexity for estimating GMM parameters \cite{vempala2004spectral,arora2005learning,chaudhuri2008learning,dasgupta2007probabilistic,moitra2010settling,hsu2013learning,belkin2010polynomial}. 
Despite statistical guarantees, these methods are not computationally efficient enough for many large-scale problems. Moreover, these results assume that the data is generated from an exact generative model which never happens in reality.
In contrast, methods based on solving maximum likelihood estimation (MLE)  problem are very
popular due to computational efficiency and robustness of MLE against perturbations of the generative model \cite{donoho1988automatic}. Although MLE-based methods are popular in practice, the theory behind their optimization algorithms (such as EM method) is little understood. Most existing algorithms with theoretical performance guarantees are not scalable to the modern applications of massive size. This is mainly due to the combinatorial and non-convex nature of the underlying optimization problems. 
  
Recent advances in the fields of non-convex optimization has led to a better understandings of the mixture model inference algorithms such as EM algrithm. For example, \cite{balakrishnan2017statistical} proves that under proper initialization, EM algorithm exponentially converges to the ground truth parameters. However, no computationally efficient initialization approach is provided. \cite{xu2016global} %
globally analyzes  EM algorithm applied to the mixture of two equally weighted Gaussian distributions. While \cite{daskalakis2016ten} provides global convergence guarantees for the EM algorithm, \cite{jin2016local} studies the landscape of GMM likelihood function with more than 3 components and shows that there might be some spurious locals even for the simple case of the equally weighted GMM.

In this work, we revisit the EM algorithm under Laplacian mixture model and Gaussian mixture model. We first show that, similar to the Gaussian case, the maximum likelihood estimation objective has no spurious local optima in the symmetric  Laplacian mixture model (LMM) with $K=2$ components. This Laplacian mixture structure has wide range of applications in medical image denoising, video retrieval and blind source separation \cite{bhowmick2006laplace, klein2014fisher,mitianoudis2005overcomplete, amin2007application, rabbani2009wavelet}.  
For the case of mixture model with $K \geq 3$ components, we propose a stochastic algorithm which utilizes the likelihood function as well as moment information of the mixture model distribution. Our numerical experiments show that our algorithm outperforms the Na\"{i}ve EM algorithm in almost all scenarios.

\section{Problem Formulation}
The general mixture model distribution is defined as     
\begin{equation*}
P(\textbf{x}; \textbf{w}, K,  \boldsymbol{\theta}) = \sum_{k = 1}^{K} w_{k}f(\textbf{x}; \boldsymbol{\theta}_{k})
\end{equation*}
where $K$ is the number of mixture components;  $\textbf{w} = (w_{1}, w_{2},..., w_{K})$  is the non-negative mixing weight with  $\sum_{k = 1}^{ K} w_{k} = 1$ and $\boldsymbol{\theta}$ = $(\boldsymbol{\theta}_{1}, \boldsymbol{\theta}_{2},..., \boldsymbol{\theta}_{K})$  is the distribution's parameter vector. Estimating the parameters of the mixture models $(\textbf{w}, \boldsymbol{\theta}, K)$ is central in many applications. This estimation is typically done by solving MLE problem due to its intuitive justification and its robust behavior \cite{donoho1988automatic}. 

The focus of our work is on the population likelihood maximization, i.e., when the number of samples is very large. When parameters $\textbf{w}$ and $K$ are known, using the law of large numbers, MLE problem leads to the following \textit{population risk} optimization problem \cite{xu2016global, daskalakis2016ten, jin2016local}:
\begin{equation}\label{E_Eq30}
\boldsymbol{\theta^{*}} = \arg\max_{\boldsymbol{\theta}} \;\;\mathbb{E} \Bigg[\log \ 
\Big(\sum _{k = 1} ^{ K } w_{k} f(\textbf{x};\boldsymbol{\theta}_{k})\Big)\Bigg]
\end{equation} 
In this paper, we focus on the case of equally weighted mixture components, i.e., $w_k = 1/K, \;\forall k$ \cite{xu2016global, daskalakis2016ten, jin2016local, srebro2007there}. We also restrict ourselves to two widely-used Gaussian mixture models and Laplacian mixture models \cite{bhowmick2006laplace, klein2014fisher,mitianoudis2005overcomplete, amin2007application, rabbani2009wavelet, vempala2004spectral,arora2005learning,chaudhuri2008learning,dasgupta2007probabilistic}. It is worth mentioning that even in these restricted scenarios, the above MLE problem is  non-convex and highly challenging to solve.

\section{EM for the case of $K=2$}
Recently, it has been shown that the EM algorithm recovers the ground truth distributions for equally weighted Gaussian mixture model with $K=2$ components \cite{xu2016global, daskalakis2016ten}. Here we extend this result to single dimensional Laplacian mixture models.

Define the Laplacian distribution with the probability density function $L(x;\mu,b) = \frac{1}{2b}e^{-\frac{|x-\mu|}{b}}$ where $\mu$ and $b$ control the mean and variance of the distribution. Thus, the equally weighted Laplacian mixture model with two components has probability density function:
\begin{equation*}
P(x;\mu_{1},\mu_{2},b) = \frac{1}{2} L(x;\mu_{1},b) + \frac{1}{2} L(x;\mu_{2},b).
\end{equation*}
In the population level estimation, the overall mean of the data, i.e., $\frac{\mu_1 + \mu_2}{2}$ can be estimated accurately. Hence, without loss of generality, we only need to estimate the normalized difference of the two means, i.e., $\mu^* \triangleq \frac{\mu_1 - \mu_2}{2}$. 
Under this generic assumption, our observations are drawn from the distribution
\begin{equation*}
P(x;\mu^*,b) = \frac{1}{2} L(x;\mu^*,b) + \frac{1}{2} L(x;-\mu^*,b).
\end{equation*}
 Our goal is to estimate the parameter $\mu^*$ from observations $x$ at the population level.
Without loss of generality, and for simplicity of the presentation, we set $b=1$, define $p_{\mu}(x) \triangleq P(x;\mu,1)$ and $L(x;\mu) \triangleq L(x;\mu,1)$. Thus, the $t$-th step of the EM algorithm for estimating the  ground truth parameter $\mu^*$ is:
%
%
%
%
\begin{equation}
\begin{aligned}\label{EMLaplacianIterate}
\lambda^{t+1} =  \frac{E_{x \sim p_{\mu^*}} \left[x\frac{0.5 L(x;\lambda^{t})}{p_{\lambda^{t}}(x)}\right]}{E_{x \sim p_{\mu}^*} \left[\frac{0.5 L(x;\lambda^{t})}{p_{\lambda^{t}}(x)}\right]},
\end{aligned}
\end{equation}
 where $\lambda^{t}$ is the estimation of $\mu^*$ in $t$-th iteration; see \cite{daskalakis2016ten,xu2016global,jin2016local} for the similar Gaussian case. 
In the rest of the paper, without loss of generality, we assume that $ \lambda^0,\mu^*> 0$.  Further, to simplify our analysis, we define the mapping  
\[
M(\lambda, \mu) \triangleq \frac{E_{x \sim p_{\mu}} \left[x\frac{0.5 L(x;\lambda)}{p_{\lambda}(x)}\right]}{E_{x \sim p_{\mu}} \left[\frac{0.5 L(x;\lambda)}{p_{\lambda}(x)}\right]}.
\]
It is easy to verify that $M(\mu^*,\mu^*) = \mu^*$, $M(-\mu^*,-\mu^*) = -\mu^*$, $M(0,0) = 0$, and $\lambda^{t+1} = M(\lambda^t,\mu^*)$. In other words, $\lambda \in \{ \mu^*,-\mu^*, 0\}$ are the fixed points of the EM algorithm. 
Using symmetry, we can simplify  $M(\cdot, \cdot)$ as
%

\begin{equation}
\begin{aligned}
M(\lambda, \mu) 
& =  E_{x \sim L(x;\mu)}\left[ x\frac{ L(x;\lambda ) -  L(x;-\lambda )}{ L(x;\lambda ) +  L(x;-\lambda )} \right]
\end{aligned}
\end{equation}
Let us first establish few lemmas on the behavior of the mapping $M(\cdot,\cdot)$.
 
%
\begin{lemma}\label{lemma220} 
The derivative of the mapping $M(\cdot)$ with respect to $\lambda$ is positive, i.e., 
	$0 < \frac{\partial }{\partial \lambda} M(\lambda,\mu)$.
\end{lemma}
\begin{proof*}
First notice that  $\frac{\partial}{\partial \lambda} M(\lambda,\mu)$ is equal to
\[
	 E_{x \sim L(x;\mu)} \left[2 x \frac{( \textrm{sign}(x - \lambda) + \textrm{sign}(x + \lambda) )(e^{-|x-\lambda| - |x+\lambda|}) ) }{( e^{-|x-\lambda|} + e^{-|x+\lambda|} ) ^{2}} \right].
\]
We prove the lemma for the following two different cases separately:
	
\noindent Case 1) $\mu < \lambda$:
	\begin{equation} \nonumber
	\begin{aligned}
	& \frac{\partial M}{\partial \lambda}
	= \frac{2}{(e^{\lambda}+e^{-\lambda})^{2}} \Big[e^{-\mu}\int_{-\infty}^{-\lambda} xe^{x} \diff x + 
	e^{\mu} \int_{\lambda}^{\infty} xe^{-x} \diff x \Big]\\
	 =& 2 \frac{(\lambda + 1) e^{-\lambda} (e^{-\mu} + e^{\mu}) }{(e^{-\lambda} + e^{\lambda})^{2}} = \frac{(\lambda + 1) e^{-\lambda} (\cosh(\mu)) }{\cosh(\lambda)^{2}} > 0. \\
	\end{aligned}
	\end{equation}
	
\noindent Case 2) $\mu > \lambda$ 
	\begin{equation}\nonumber
	\begin{aligned}
	& \frac{\partial M}{\partial \lambda} 
	=  \frac{2e^{-\mu}\Big[\int_{-\infty}^{-\lambda} xe^{x} \diff x +   \int_{\lambda}^{\mu} xe^{x} \diff x + e^{2\mu} \int_{\mu}^{\infty} xe^{-x} \diff x \Big]}{(e^{\lambda}+e^{-\lambda})^{2}}  \\
	 = &	\frac{2}{(e^{\lambda}+e^{-\lambda})^{2}} \Big[e^{-\mu}\Big((\lambda+1)e^{-\lambda}-(\lambda-1)e^{\lambda}\Big) +2\mu\Big] \\
	 \geq  & \frac{2}{(e^{\lambda}+e^{-\lambda})^{2}} \Big[e^{-\mu}\Big((\mu+1)e^{-\mu}-(\mu-1)e^{\mu}\Big) +2\mu\Big] >0.\; \QED   
	\end{aligned}
	\end{equation}
	\end{proof*}
\begin{lemma}\label{lemma2}
	For $  0<\lambda < \eta  $, we have
	\begin{equation} \nonumber
	\begin{aligned}
	\frac{\partial}{\partial \eta} M(\lambda,\eta) =  1- 2\frac{e^{-\eta}\lambda + e^{-\lambda}}{e^{\lambda}+e^{-\lambda}} > 1 - 2\frac{e^{-\lambda}\lambda + e^{-\lambda}}{e^{\lambda}+e^{-\lambda}} >0 
	\end{aligned}
	\end{equation}	
\end{lemma}
\begin{proof*}
	When $ \eta> \lambda$, it is not hard to show that
$
  M(\lambda, \eta) = \frac{1}{2} e^{-\eta}\Big\{  \tanh(\lambda)(\lambda + 1)e^{-\lambda} + (\lambda - 1)e^{\lambda} +  (\lambda + 1)e^{-\lambda}  - (\lambda - 1)e^{\lambda} \tanh(\lambda) \Big\}+ \tanh(\lambda)\eta.$
	Hence,
	\begin{equation} \nonumber
	\begin{aligned}
	& \frac{\partial M}{\partial \eta} 
	 = -e^{-\eta}\left\{\frac{\lambda + 1}{e^{\lambda}+e^{-\lambda}}  
	+\frac{\lambda - 1}{e^{\lambda}+e^{-\lambda}} \right\}  + \tanh(\lambda) \\
	= \;& \frac{-2e^{-\eta}\lambda + e^{\lambda}-2e^{-\lambda} + e^{-\lambda} }{e^{\lambda}+e^{-\lambda}}  
	> 1 - 2 \frac{\lambda e^{-\lambda} + e^{-\lambda}}{e^{\lambda} + e^{-\lambda}} > 0,
	\end{aligned}
	\end{equation} 
	where the last two inequalities are due to the facts that $\lambda <\eta$ and $\frac{\lambda e^{-\lambda} + e^{-\lambda}}{e^{\lambda} + e^{-\lambda}} <1/2$.  \QED 
\end{proof*}
\newpage
\begin{theorem} \label{TH1}
	 Without loss of generality, assume that $\lambda^0,\mu^*>0$. Then the EM iterate defined in \eqref{EMLaplacianIterate} is a contraction, i.e., 
$
	\bigg|\frac{\lambda^{t+1}-\mu^*}{\lambda^{t}-\mu^* } \bigg|< \kappa < 1, \;\forall t,
$
	where  $\kappa = \max \ \{\kappa_1,\kappa_{2}\}$, $\kappa_{1} = \frac{(\mu^* + 1) e^{-\mu^*}  }{cosh(\mu^*)} $,  and $\kappa_{2} =  2 \frac{\lambda^{0} e^{-\lambda^{0}} + e^{-\lambda^{0}}}{e^{\lambda^{0}} + e^{-\lambda^{0}}}$.
\end{theorem}
Theorem~\ref{TH1} shows that the EM iterates converge to the ground truth parameter which is the global optimum of the MLE. 
\begin{proof*}
First of all, according to the Mean Value Theorem,  
	$\exists \; \xi$ between $\lambda^{t}$ and $\mu^*$ such that:
	\begin{small}
	\begin{equation} \nonumber
	\begin{aligned}
	\frac{\lambda^{t+1} - \mu^*}{\lambda^{t} - \mu^*} = \frac{M(\lambda^{t}, \mu^*)  - M(\mu^*, \mu^*)}{\lambda^{t}- \mu^*} = \frac{\partial }{\partial \lambda} M(\lambda,\mu^*)\Big|_{\lambda = \xi} > 0, \\
	\end{aligned}
	\end{equation}
	\end{small}
	
\noindent where the inequality is due to lemma ~\ref{lemma220}. Thus, $\lambda^{t}$ does not change sign during the algorithm. 	
	  Consider two different regions: $\mu^* > \lambda$ and $\mu^* < \lambda$. When $\mu^* < \lambda$, case 1 in Lemma~\ref{lemma220} implies that
	 \begin{small} 
	\begin{equation} \nonumber
	\begin{aligned}
	\frac{\partial M}{\partial \lambda} \Big|_{\lambda = \xi} & = \frac{(\xi + 1) e^{-\xi} (\cosh(\mu^*)) }{\cosh(\xi)^{2}}\leq  \frac{(\mu^* + 1) e^{-\mu^*}  }{\cosh(\mu^*)}  = \kappa_{1} < 1.
	\end{aligned}
	\end{equation}
	\end{small}

\noindent The last two inequalities are due to the fact that $\mu^*<\xi<\lambda $, and the fact that $f(\xi) = \frac{(\xi + 1) e^{-\xi}  }{cosh(\xi)^{2}}$ is a positive and decreasing function in $\mathbb{R}^{+}$ with $f(0) = 1$.
	On the other hand, when $\mu^* > \lambda$, the Mean Value Theorem implies that
	 \begin{small} 
	\begin{equation} \nonumber
	\begin{aligned}
	&\frac{\lambda^{t+1} - \mu^* }{\lambda^{t} - \mu^*} 
	 = \frac{\lambda^{t+1} - \lambda^{t} }{\lambda^{t} - \mu^*}+1 
	 =  \frac{M(\lambda^{t},\mu^*) - M(\lambda^{t},\lambda^{t})}{\lambda^{t} - \mu^*} + 1  \\
	 = \;&1 - \frac{\partial }{\partial \mu^*} M(\lambda^{t},\mu^*)\Big|_{\mu^* = \eta} \leq 2 \frac{\lambda^{t} e^{-\lambda^{t}} + e^{-\lambda^{t}}}{e^{\lambda^{t}} + e^{-\lambda^{t}}} \\ 
	 \leq \;& 2 \frac{\lambda^{0} e^{-\lambda^{0}} + e^{-\lambda^{0}}}{e^{\lambda^{0}} + e^{-\lambda^{0}}}  =\kappa_{2} < 1,
	\end{aligned}
	\end{equation}
	\end{small}
	
\noindent where the last two inequalities are due to lemma ~\ref{lemma2} and the facts that 1) $\lambda^{t}$ does not change sign and 2) the function $f(\lambda) = 2 \frac{\lambda e^{-\lambda} + e^{-\lambda}}{e^{\lambda} + e^{-\lambda}} $ is  positive and decreasing  in $\mathbb{R}^{+}$ with $f(0) = 1$. Hence,
$	\frac{\lambda^{t+1} - \mu^* }{\lambda^{t} - \mu^*} < \kappa_{2} < 1. 
$	Combining the above two cases will complete the proof. \QED
\end{proof*}

\section{Modified EM for the case of $K\geq 3$}
\label{Section 4}
In \cite{srebro2007there} it is conjectured that the local optima of the population level MLE problem for any  equally weighed GMM is globally optimal. Recently,  \cite{jin2016local} has rejected this conjecture by providing a counter example with $ K = 3$ components. Moreover, they have shown that the local optima could be arbitrary far from ground truth parameters and there is a positive probability for the EM algorithm with random initialization to converge to these spurious local optima. Motivated by \cite{jin2016local}, we numerically study the performance of the  EM algorithm in both GMMs and  LMMs.


\begin{figure}
	\centering
	\includegraphics[width=.8\linewidth, height=.6\linewidth]{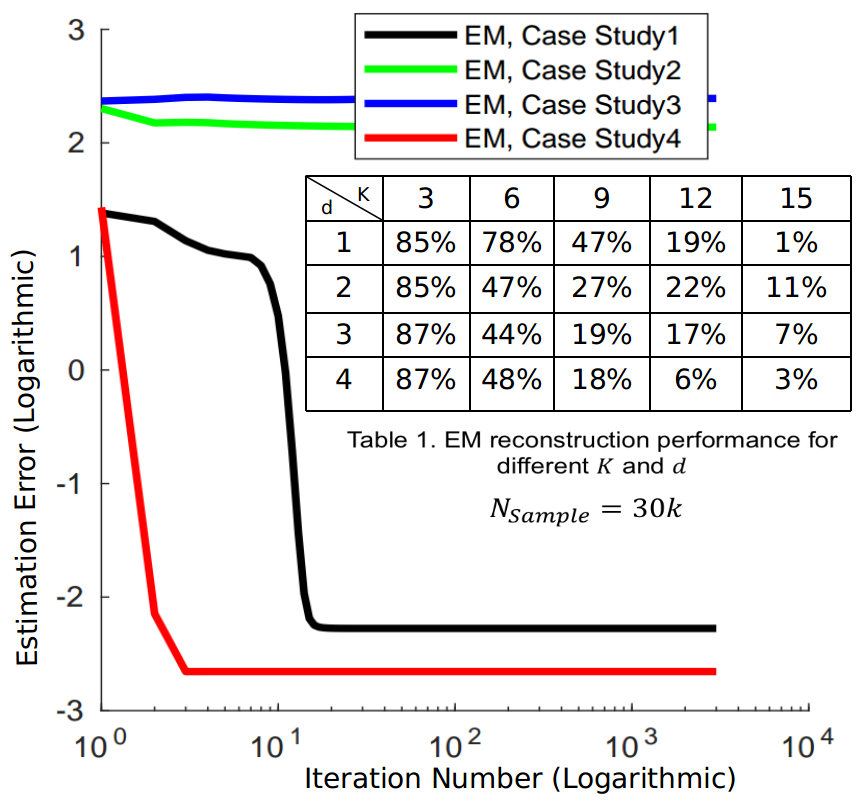}
	\caption{Na\"{i}ve EM  fails to recover the ground truth parameter.}
	\label{fig:Naive}
\end{figure}
\noindent\textbf{Numerical Experiment 1}:
Figure~\ref{fig:Naive} presents the convergence plots of the EM algorithm with four different initializations. Two of these initializations converge to the global optima, while the other two fails to recover the ground truth parameters and they are trapped in spurious local optima. To understand the performance of the EM algorithm with random initialization, we ran the EM algorithm for different number of components~$K$ and dimensions~$d$. First we generate the $d$-dimensional mean vectors $\bmu_k\sim N(\textbf{0}, 5 \mathbf{I})$, $k=1, \ldots,K$. These vectors are the mean values of different Gaussian components. For each Gaussian component, the variance is set to~$1$. Thus the vectors $\bmu_1,\bmu_2,\ldots, \bmu_K$ will completely characterize the distribution of the GMM. Then, $30,000$ samples are randomly drawn from the generated GMM and the EM algorithm is run with 1000 different initializations, each for   $3000$ iterations. The table in Figure~\ref{fig:Naive} shows the percentage of the times that the EM algorithm converges to the ground truth parameters (global optimal point) for different values of $K$ and $d$. As can be seen in this table, EM fails dramatically especially for larger values of $K$. 
%

By examining the spurious local optima in the previous numerical experiment, we have noticed that many of these local optima fail to satisfy the first moment condition. More specifically, we know that 
any global optimum of the MLE problem~\eqref{E_Eq30} should recover the ground truth parameter -- up to permutations \cite{teicher1963identifiability,allman2009identifiability}. Hence, any global optimum~$\hat{\bmu} = (\hat{\bmu}_1,\ldots,\hat{\bmu}_K)$ has to satisfy the first moment condition
$
\mathbb{E}(\textbf{x}) = \sum_{k = 1}^{K}\frac{1}{K}\hat{\bmu}_{k}.
$
Without loss of generality and by shifting all data points, we can assume that $\mathbb{E}(\textbf{x}) = 0$. Thus, $\hat{\bmu}$ must satisfy  the condition  
 \begin{equation} \label{EqMC}
 \sum_{k = 1}^{K}\hat{\bmu}_{k} = 0.
 \end{equation}
However, according to our numerical experiments, many spurious local optima fail to satisfy \eqref{EqMC}.
To enforce  condition~\eqref{EqMC}, one can regularize the MLE cost function with the first order moment condition and solve
\begin{equation}  \label{eq:ModifiedGMMK3}
	\begin{aligned}
		\max_{\boldsymbol{\mu}} \;
		  \mathbb{E}_{\bmu^*}\left[ \log  
		\left( \sum_{k = 1} ^{K} \frac{1}{K} f(\textbf{x};\boldsymbol{\mu}_{k}) \right)\right] - \frac{M}{2} \norm{\sum_{k = 1}^{K} \boldsymbol{\mu}_{k}}_{2}^2,
	\end{aligned}
\end{equation}
where $M >0$ is the regularization coefficient. To solve~\eqref{eq:ModifiedGMMK3}, we propose the following iterative algorithm:
\small
\begin{equation} \label{eq:ModifiedEMAlgo}
\begin{aligned}
&\boldsymbol{{\mu}}_k^{t+1} = \frac{\mathbb{E}_{\bmu^*} [\textbf{x} w^t_k(\textbf{x})] + M K \bmu_{k}^{t}- M\sum_{j = 1}^{K} \boldsymbol{\mu}_j^t}{ M K+ \mathbb{E}_{\boldsymbol{\mu^*}}[w_k^t(\textbf{x})]},\;\forall k,
\end{aligned}
\end{equation}
\normalsize
where  $w_k^t(\textbf{x}) \triangleq \frac{ f(\textbf{x}; \boldsymbol{\mu}_{k}^{t})}{\sum_{j=1}^K f(\textbf{x}; \boldsymbol{\mu}_{j}^{t}) }, \forall k = 1,\ldots,K$.  This algorithm is based on the successive upper-bound minimization framework  \cite{razaviyayn2014parallel,razaviyayn2013unified,hong2016unified}.  Notice that if we set $M=0$ in \eqref{eq:ModifiedEMAlgo}, we obtain the na\"ive EM algorithm. The following theorem establishes the convergence of the iterates in \eqref{eq:ModifiedEMAlgo}.
\begin{theorem}
	Any limit point of the iterates generated by \eqref{eq:ModifiedEMAlgo}  is a stationary point of \eqref{eq:ModifiedGMMK3}.
\end{theorem}
\noindent\textbf{Proof Sketch} 
Let $g(\bmu)$ be the objective function of ~\eqref{eq:ModifiedGMMK3}. Using Cauchy-Schwarz and Jensen's inequality, one can show that
\small $g(\bmu) \geq  \widehat{g}(\bmu, \bmu^t) \triangleq\mathbb{E}_{\bmu^*}\left[ \sum_k\left(w_{k}^t(\textbf{x}) \log  
 \left(\frac{f(\textbf{x};\boldsymbol{\mu}_{k})}{ f(\bx;\bmu_k^t)}\right) \right)\right] - \frac{M}{2} K \sum_k\norm{ \boldsymbol{\mu}_{k}  -  \boldsymbol{\mu}_{k}^{t} }_{2}^2 - M \langle \sum_k(\bmu_k - \bmu_k^t),\sum_k\bmu_{k}^{t}\rangle + g(\bmu^t) $. 
 \normalsize Moreover,  $\widehat{g}(\bmu^t,\bmu^t) = g(\bmu^t)$. Thus the assumptions of \cite[Proposition 1]{razaviyayn2013unified}  are satisfied. Furthermore, notice that the iterate~\eqref{eq:ModifiedEMAlgo} is obtained based on the update rule~$\bmu^{t+1} = \arg\min_{\bmu} \widehat{g}(\bmu, \bmu^t)$. Therefore,  \cite[Theorem 1]{razaviyayn2013unified} implies that every limit point of the algorithm is a stationary point.  \QED
\\
 
\begin{figure}
\vspace{-0.cm}
	\centering
	\includegraphics[width=.8\linewidth, height=.25\linewidth]{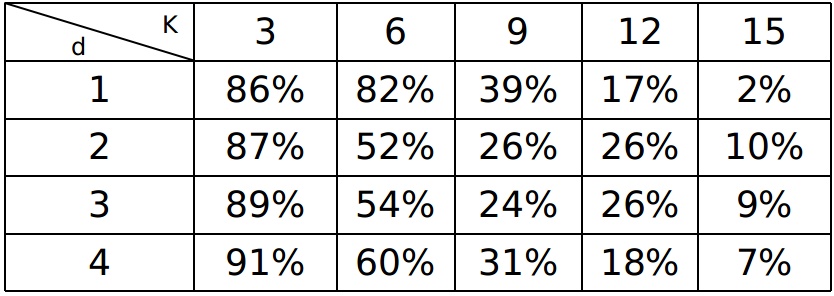}
	\caption{Modified EM based on regularized MLE~\eqref{eq:ModifiedEMAlgo}.}
	\label{fig:NaiveVSModified}
\vspace{-0.5cm}
\end{figure}

\noindent \textbf{Numerical Experiment 2}: To evaluate the performance of the algorithm defined in~\eqref{eq:ModifiedEMAlgo}, we repeat the Numerical Experiment~1 with the proposed iterative method \eqref{eq:ModifiedEMAlgo}. 
Figure \ref{fig:NaiveVSModified} shows the performance of the proposed iterative method \eqref{eq:ModifiedEMAlgo}. 
As can be seen from the table, the regularized method still fails to recover the ground truth parameters in many scenarios.  More specifically, although the regularization term enforces \eqref{EqMC}, it changes the likelihood landscape and hence, it introduces some new spurious local optima.\\

In our numerical experiment~2, we observed that many of the spurious local optima are tied to a fixed value of $M$. In other words, after getting stuck in a spurious local optimum point, changing the value of $M$ helps us escape from that local optimum. Notice that the global optimal parameter~$\bmu^*$ is the solution of~\eqref{eq:ModifiedGMMK3} for any value of $M$. Motivated by this observation, we consider the following  objective function:

\small
\begin{equation} \label{EQS}
	\begin{aligned}
\max_{\boldsymbol{\mu}} \; \mathbb{E}_{\lambda \sim \Lambda} \Bigg[\mathbb{E}_{\bmu^*}\left[ \log  
\left( \sum_{k = 1} ^{K} \frac{1}{K} f(\textbf{x};\boldsymbol{\mu}_{k}) \right)\right] 
	 -\frac{\lambda}{2} \norm{\sum_{k = 1}^{K}\boldsymbol{\mu}_{k}}_{2}^2 \Bigg], 
	\end{aligned}
\end{equation}
\normalsize
where $\Lambda$ is some continuous distribution defined over $\lambda$.  The idea behind this objective is that each sampled value of $\lambda$ leads to different set of spurious local optima. However, if a point $\widehat{\bmu}$ is a fixed point of EM algorithm for any value of $\lambda$, it must be a stationary point of the MLE function and also it should satisfy the first moment  condition~\eqref{EqMC}.  Based on this objective function, we propose algorithm~\ref{alg:StochasticBSUMAlg} for estimating the ground truth parameter.

\begin{algorithm} 
	\caption{Stochastic multi-objective EM}
	\label{alg:StochasticBSUMAlg}
    \textbf{Input:} {Number of iterations: $N_{Itr}$,  distribution $\Lambda$,  Initial estimate: $\boldsymbol{\mu^{0}}$ }\\
	\textbf{Output:} {$\boldsymbol{\hat{\mu}}$}\\
	\For{$t =1:N_{Itr} $}
	{
			  Sample $\;\lambda \sim \Lambda,$

		\For{$k =1:K $}
		{

	$w_k^{t-1}(\textbf{x}) \triangleq \frac{ f(\textbf{x}; \boldsymbol{\mu}_{k}^{t-1})}{\sum_{j=1}^K f(\textbf{x}; \boldsymbol{\mu}_{j}^{t-1}) };$
	
	$\boldsymbol{{\mu}}_k^{t} = \frac{\mathbb{E}_{\bmu^*} [\textbf{x} w^{t-1}_k(\textbf{x})] + \lambda K \bmu_{k}^{t-1}- \lambda\sum_{j = 1}^{K} \boldsymbol{\mu}_j^{t-1}}{ \lambda K+ \mathbb{E}_{\boldsymbol{\mu^*}}[w_k^{t-1}(\textbf{x})]};$\

		}
	$\boldsymbol{\hat{\mu}^{t}} = (\hat{\bmu}_1^{t},\ldots,\hat{\bmu}_K^{t})$
	}
	{
		return $\boldsymbol{\hat{\mu}}^{N_{Itr}}$
	}

\end{algorithm}

\begin{figure}
		\centering
		\includegraphics[width=.8\linewidth, height=.25\linewidth]{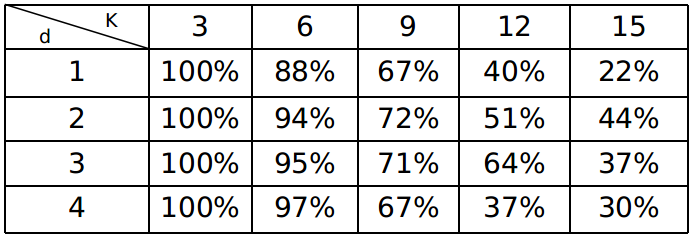}
		\caption{Performance of Stochastic multi-objective EM. 
		}
	\label{fig:Multi}
\end{figure}

%
%
%
%
%
\noindent\textbf{Numerical Experiment 3}:
To evaluate the performance of Algorithm 1, we repeat the data generating procedure in Numerical Experiment~1. Then we run Algorithm 1 on the generated data. Figure ~\ref{fig:Multi} shows the performance of this algorithm. As can be seen from this figure, the proposed method significantly improves the percentage of times that a random initialization converges to the ground truth parameter.  For example, the proposed method converges to the global optimal parameter $70\%$ of the times for $K=9, d = 3$, while the na\"ive EM  converges for $19\%$ of the initializations (comparing Fig.~\ref{fig:Naive} and Fig.~\ref{fig:Multi}).\\


\noindent\textbf{Remark:} While the results in this section are only presented for GMM model, we have observed similar results in LMM model. These results are omitted due to lack of space.

\section{Conclusion}
In this paper, first the convergence behavior of the EM algorithm for equally weighted Laplacian mixture model with two components is studied. It is shown that the EM algorithm with random initialization converges to the ground truth distribution with probability one. Moreover, the landscape of the equally weighted mixture models with more than two components is revisited. Based on our numerical experiments, we proposed a modified EM approach which significantly improves the probability of recovering the ground truth parameters.

%
%


\newpage
\small
\bibliographystyle{IEEEtran}
\bibliography{mybib}

\end{document}